\documentclass[10pt,letterpaper]{article}

\usepackage[preprint,nonatbib]{customize}
\usepackage[T1]{fontenc}

\usepackage{microtype}

\usepackage[authoryear,round]{natbib}
\setcitestyle{citesep={;},aysep={,},yysep={;}}

\usepackage{graphicx}
\usepackage{amsmath}
\usepackage{amssymb}
\usepackage{array}
\usepackage{booktabs}
\usepackage{tabularx}
\usepackage{float}
\usepackage{placeins}
\usepackage{caption}
\usepackage{listings}
\usepackage{xcolor}
\usepackage{mdframed}
\usepackage{needspace}

\lstdefinestyle{toolschema}{
  basicstyle=\ttfamily\scriptsize,
  frame=single,
  framerule=0.4pt,
  breaklines=true,
  breakatwhitespace=true,
  columns=fullflexible,
  keepspaces=true,
  showstringspaces=false,
  aboveskip=4pt,
  belowskip=4pt,
  captionpos=b
}

\usepackage{amsmath,amsfonts,bm}

\def\eqref#1{equation~\ref{#1}}

\def\1{\bm{1}}

\DeclareMathAlphabet{\mathsfit}{\encodingdefault}{\sfdefault}{m}{sl}
\SetMathAlphabet{\mathsfit}{bold}{\encodingdefault}{\sfdefault}{bx}{n}

\usepackage[hidelinks,pdfusetitle]{hyperref}
\usepackage{url}

\title{ClueFlow: Learning Long-Horizon Search through Attributed Research-State Transitions}

\author{
  Yicheng Wang\thanks{Work done during an internship at Joy Future Academy.
  Email: \texttt{yichengwang.992@gmail.com}.} \quad
  Wentao Tan \quad
  Yidan Huang \quad
  Yifei Liu \quad
  Qiong Cao\thanks{Corresponding author:
  \texttt{mathqiong2012@gmail.com}.} \\[0.5em]
  Joy Future Academy \\[0.6em]
  {\small
    \href{https://github.com/clueflow/clueflow}{%
      \raisebox{-0.15em}{%
        \includegraphics[height=1em]{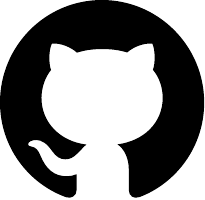}%
      }\hspace{0.4em}GitHub%
    }\qquad
    \href{https://huggingface.co/collections/wangyicheng992/clueflow}{%
      \raisebox{-0.2em}{%
        \includegraphics[height=1.1em]{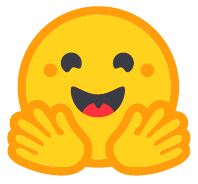}%
      }\hspace{0.4em}Hugging Face%
    }
  }
}
\hypersetup{
  pdfauthor={Yicheng Wang, Wentao Tan, Yidan Huang, Yifei Liu, Qiong Cao}
}
\date{}

\begin{document}
\raggedbottom
\setlength{\parskip}{4pt}

\maketitle
\setcounter{footnote}{0}

\vspace{-2em}
\begin{abstract}
Long-horizon search requires agents to acquire relevant evidence and integrate
it into an evolving research state. Existing methods maintain research states
to guide exploration or assign turn-level progress signals, yet grounding these
signals in explicit, attributable state changes remains underexplored. We
introduce ClueFlow, which learns long-horizon search from attributed
research-state transitions. ClueFlow constructs Source--State--Final
attribution chains that connect evidence-acquiring browser actions, Evidence
and Candidate updates, and the final answer supported by those updates. During
search, ClueFlow records typed state changes and potentially supporting browser
actions, then applies gold-conditioned semantic validation and source tracing
after execution. We use this attribution to guide
both supervised fine-tuning and reinforcement learning. During SFT, attribution
selects which decisions receive supervision with their original interaction
contexts. During Turn-PPO, the same attribution augments
outcome-based turn advantages with credit assigned along the
Source--State--Final chains. Transition-level validation also recovers useful
supervision from gold-consistent state changes in trajectories with incorrect
final answers. We evaluate ClueFlow across two Qwen backbones and four search
benchmarks spanning live-web and fixed-corpus settings, observing consistent
gains from SFT and further improvements from PPO. With Qwen3.5-35B-A3B,
ClueFlow achieves 60.0\% accuracy on BrowseComp and 72.0\% on BrowseComp-Plus.
In controlled BrowseComp-Plus experiments, attribution-guided SFT
matches all-turn exact-context SFT using about 30\% of the training tokens,
while attribution-guided PPO outperforms outcome-only PPO by 4.1~pp.
These results demonstrate that state-transition attribution enables efficient
supervision and effective credit assignment for long-horizon search agents.
\end{abstract}

\section{Introduction}

Long-horizon search requires agents to gather evidence, evaluate candidate
answers, and revise their hypotheses across extended interactions. For a
multi-constraint question, a page may reveal evidence that confirms or rules out
a candidate, prompting the agent to preserve that evidence and update its
research state. A correct final
answer alone does not distinguish useful evidence acquisition and state updates
from routine exploration. Conversely, filtering out trajectories with incorrect
final answers may discard valid intermediate progress.
Learning from long search trajectories therefore requires identifying which
research-state changes constitute meaningful progress and which
information-gathering actions support those changes.

\begin{figure}[t]
    \centering
    \includegraphics[width=\linewidth]{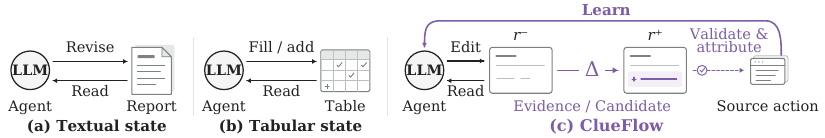}
    \caption{Learning from attributed research-state transitions. (a)
    Textual-state agents maintain evolving reports that summarize accumulated
    findings (e.g., IterResearch; \citealp{chen2026iterresearch}). (b)
    Tabular-state agents maintain tables tracking candidates, constraints, and
    collected information (e.g., Table-as-Search;
    \citealp{lan2026tableassearch}). Both update and consult these states to guide
    subsequent search. (c) ClueFlow additionally learns from record-level
    transitions through post-execution validation and source attribution.}
    \label{fig:state-comparison}
\end{figure}

Recent methods organize accumulated findings into explicit research states, such
as evolving reports~\citep{chen2026iterresearch}, structured
summaries~\citep{zhu2026retrac}, and candidate--constraint
tables~\citep{lan2026tableassearch}, to preserve useful information and guide
subsequent exploration.
Related memory methods learn to manage retained information through outcome-based
reinforcement learning~\citep{yan2026memoryr1} or explicit feedback on memory
informativeness~\citep{li2026mempo}. Other methods select valuable steps for
supervision~\citep{chen2025atlas,slinko2026srft} or provide fine-grained credit
assignment~\citep{wang2025stepsearch,wang2026igpo,liu2026pica,wang2026cwgrpo}.
However, using validated research-state changes and explicit links to their
supporting search actions to guide supervision and credit assignment remains
underexplored. Figure~\ref{fig:state-comparison} illustrates this distinction.

Our key insight is that validating research-state transitions and tracing their
supporting evidence makes explicit both what progress was made and which search
actions supplied the evidence for that progress. A state records the agent's
current evidence and answer hypotheses; a transition makes explicit how these
records change. Based on this insight, ClueFlow constructs
Source--State--Final attribution chains that connect evidence-acquiring browser
actions, Evidence and Candidate transitions, and the final answer supported by
those transitions.

During search, ClueFlow maintains persistent Evidence and Candidate records and
logs typed state changes together with potentially supporting browser actions
for each Evidence update. A recorded state change does not by itself establish
useful research progress. After execution, gold-conditioned semantic validation
identifies Evidence and Candidate transitions that make useful progress toward
the reference answer. Source tracing then links validated Evidence transitions
to the browser actions that supplied their supporting information.
Gold answers and validation judges are used only for post-execution
attribution and are never exposed to the search policy at test time.
For correct trajectories, the resulting attribution connects evidence-acquiring
browser actions and validated state transitions to the final answer they support.
For trajectories ending in incorrect answers, ClueFlow retains validated
Source--State partial chains that capture gold-consistent intermediate progress.

We apply this attribution principle in both supervised fine-tuning and
reinforcement learning. During supervised fine-tuning, attribution selects which
decisions are supervised with their original interaction contexts, while other
valid turns in successful trajectories are supervised with compactly serialized
contexts. During Turn-PPO~\citep{li2026turnppo}, the same attribution
augments outcome-based turn advantages with credit assigned along the
Source--State--Final chains. Both stages also use validated Source--State
progress recovered from trajectories with incorrect final answers.

Our main contributions are:

\begin{itemize}
    \item We introduce attributed research-state transitions to represent
    verifiable research progress and link it to the browser actions that supplied
    its supporting evidence, providing an explicit basis for agent learning.

    \item We develop ClueFlow, which constructs Source--State--Final attribution
    chains through semantic validation and source tracing to guide SFT context
    selection and Turn-PPO credit assignment. Both stages also learn from
    validated partial progress in trajectories with incorrect final answers.

    \item We conduct controlled ablations on BrowseComp-Plus, showing that
    attribution-guided SFT matches all-turn exact-context SFT using about 30\%
    of the training tokens, while attribution-guided PPO with failure recovery
    outperforms outcome-only PPO by 4.1~pp. ClueFlow SFT also outperforms the
    strongest tested ReAct SFT configuration by 6.5~pp. We further evaluate
    ClueFlow with two Qwen backbones on four benchmarks spanning live-web and
    fixed-corpus search, supporting the applicability of attributed
    research-state learning across different models and search environments.
\end{itemize}

\section{Related Work}

\subsection{Context and Memory Management for Long-Horizon Agents}

Long-horizon agents operate on tasks spanning tens to hundreds of steps, where
managing the ever-growing interaction history is a central challenge. Existing
methods mainly control context size by discarding past observations,
selectively retaining information, or compressing past
interactions~\citep{deepseek2025v32,glm52026,ye2025agentfold}.

Recent search agents further organize accumulated findings into persistent,
reusable task states that capture evidence, candidate answers, and future search
plans. These approaches differ in how they represent and update research states.
RE-TRAC updates the state at trajectory boundaries by
recursively compressing each completed trajectory together with the previous
state into a structured summary, which then guides the next
rollout~\citep{zhu2026retrac}. IterResearch adopts a finer-grained mechanism,
updating an evolving report after each tool
interaction~\citep{chen2026iterresearch}. Table-as-Search instead represents
the task state as a candidate--constraint table and advances search by adding
candidate rows and filling empty cells~\citep{lan2026tableassearch}. Related
work also learns how memory should be
managed. Memory-R1 uses outcome-driven reinforcement learning to learn
operations for adding, updating, and deleting entries in external
memory~\citep{yan2026memoryr1}, whereas MemPO optimizes information summarization
and retention according to their usefulness for subsequent
decisions~\citep{li2026mempo}.

Existing approaches primarily maintain research states to guide subsequent
search. ClueFlow additionally validates state transitions, traces them to
supporting browser actions, and uses the resulting attributed changes to guide
supervision and credit assignment.

\subsection{Post-Training and Credit Assignment for Search Agents}

Whole-trajectory imitation and outcome-based rewards do not explicitly
distinguish critical, redundant, and harmful turns. Recent work therefore seeks
to identify the training value of individual turns and provide more targeted
supervision.

Existing methods estimate turn value in different ways. ATLaS uses an LLM
selector to identify critical steps in expert trajectories and computes the
training loss only on those steps~\citep{chen2025atlas}. SRFT labels each step as
useful, harmful, or unnecessary and masks harmful steps, thereby recovering
useful supervision from failed trajectories~\citep{slinko2026srft}. Beyond such
binary selection, StepSearch rewards evidence gain while penalizing redundant
retrieval~\citep{wang2025stepsearch}; IGPO and PiCA assign rewards based on
changes in the probability of producing the correct answer or eventually
completing the task~\citep{wang2026igpo,liu2026pica}; and CriticSearch and
CW-GRPO use LLM judgments to produce turn-level rewards or redistribute
trajectory-level advantages~\citep{zhang2026criticsearch,wang2026cwgrpo}. These
methods convert turn-value estimates into selective losses, rewards, or
advantage weights.

Rather than scoring turns directly, ClueFlow guides SFT and Turn-PPO using
validated, source-attributed research-state transitions.

\section{Method}
\label{sec:method}

\subsection{Overview}

\begin{figure}[t]
    \centering
    \includegraphics[width=\linewidth]{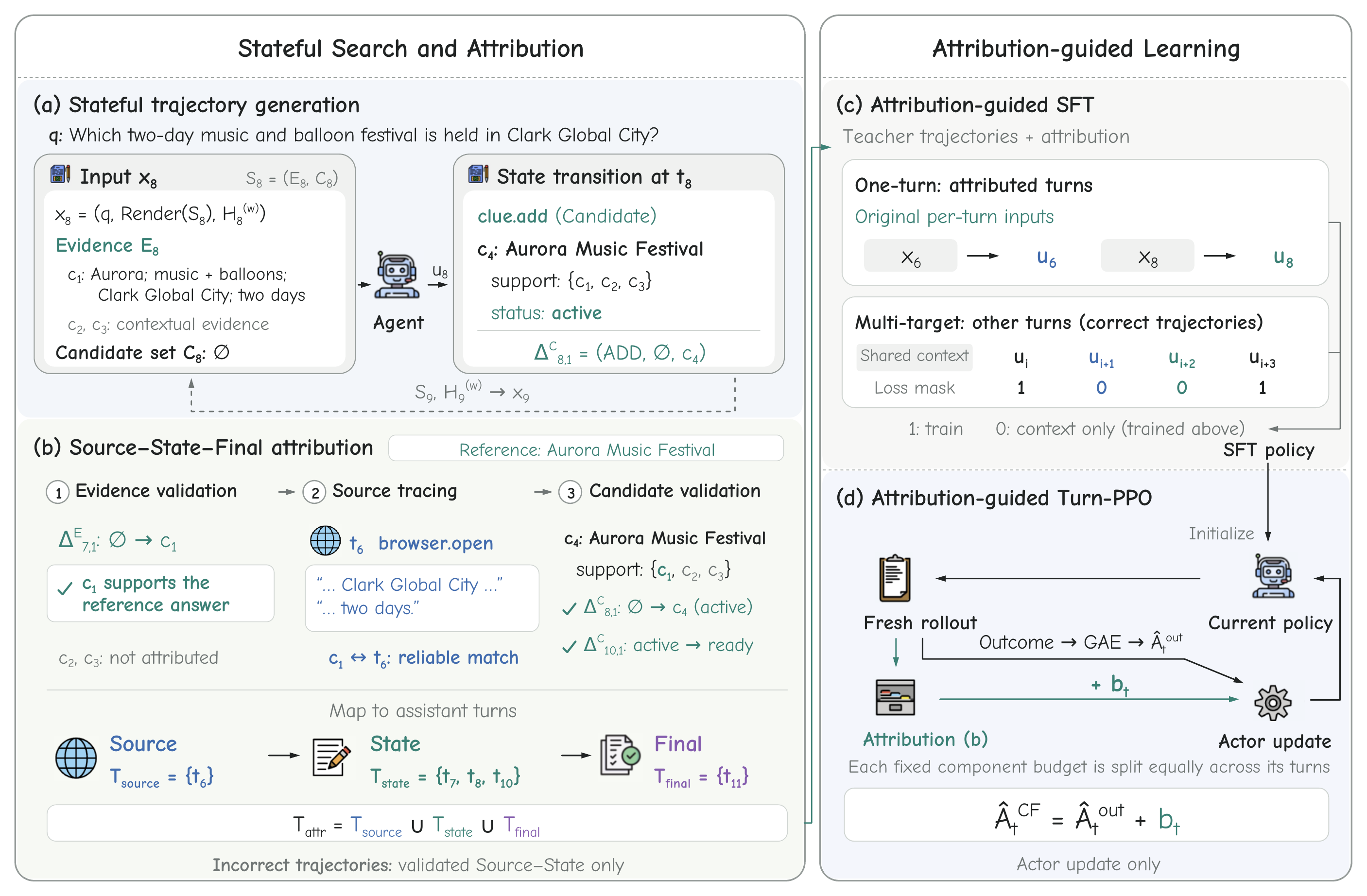}
    \caption{
    \textbf{Overview of ClueFlow.}
    \textbf{(a)} Stateful trajectory generation maintains Evidence and
    Candidate records across search turns.
    \textbf{(b)} Post-execution validation against the reference answer
    and source tracing identify attributed Source, State, and Final turns.
    Validated Source--State progress can also be retained from incorrect trajectories.
    \textbf{(c)} Attribution-guided SFT uses one-turn examples with original
    per-turn inputs for attributed turns and multi-target chunks for other
    supervised turns from correct trajectories.
    \textbf{(d)} Attribution-guided Turn-PPO adds attribution bonuses to
    outcome advantages for actor updates.
    Here, $\Delta^{E/C}_{t,k}$ denotes the $k$-th Evidence/Candidate
    state transition at assistant turn $t$.
    }
    \label{fig:method}
\end{figure}

A teacher agent interacts with the search environment to generate
long-horizon trajectories with explicit research-state transitions. ClueFlow
then validates these transitions and traces them to the browser actions that
supplied the supporting evidence. The attributed teacher trajectories are
used to construct SFT data and train an initial search policy. Starting from
this policy, PPO generates new online rollouts and applies the same attribution
procedure to assign fine-grained credit to individual turns.
Figure~\ref{fig:method} summarizes the pipeline.

\subsection{Stateful Trajectory Generation}
\label{sec:state-execution}

ClueFlow uses the same stateful search interface for teacher trajectory
generation, online policy rollouts during reinforcement learning, and test-time
inference (Figure~\ref{fig:method}(a)).

\paragraph{Research state.}
ClueFlow maintains a persistent research state for each search trajectory. We
denote the state available at the beginning of turn $t$ by
\begin{equation}
S_t=(E_t,C_t),
\qquad E_0=C_0=\varnothing,
\label{eq:research-state}
\end{equation}
where $E_t$ is the set of Evidence records containing information extracted
from browser observations, and $C_t$ is the set of Candidate records that
maintain answer hypotheses, their supporting Evidence, and their verification
progress. Each
Candidate has a status of \texttt{active},
\texttt{ready}, or \texttt{rejected}, indicating that it remains under
investigation, is eligible for submission, or has been ruled out, respectively.
Candidate records also track satisfied and
unresolved constraints, contradictions, and the next search step. The state
fields are summarized in Appendix~\ref{app:schema}.

\paragraph{Trajectory generation.}
Given a question $q$, the agent generates a search trajectory by repeatedly
producing an assistant response and executing its requested tools. It can call
\texttt{browser.search}, \texttt{browser.open}, and \texttt{browser.find} to
search the web, read a page, and locate relevant passages. It can also use
\texttt{clue.add} and \texttt{clue.update} to create or revise Evidence and
Candidate records, and \texttt{clue.discard} to remove obsolete Evidence. The
tool interfaces are summarized in Appendix~\ref{app:schema}.

At turn $t$, we construct
\begin{equation}
x_t=
\left(q,\operatorname{Render}(S_t),H_t^{(w)}\right),
\qquad
u_t\sim\pi_\theta(\cdot\mid x_t),
\label{eq:actor-input}
\end{equation}
where $x_t$ is the model input, $\operatorname{Render}(S_t)$ converts the
structured state into text, and $H_t^{(w)}$ contains up to $w$ most recent
turns preceding turn $t$. The agent model $\pi_\theta$ produces the response
$u_t$, which may contain reasoning and multiple tool calls. Fixed system
instructions and tool schemas are omitted from the notation.

The environment executes the tool calls in $u_t$: Browser results are appended
to the interaction history, while a state transition occurs whenever a
successful Clue operation modifies an Evidence or Candidate record. A single
turn may contain multiple state-changing operations. We record the $k$-th
transition in turn $t$ by its operation and the affected record's contents
before and after the change:
\begin{equation}
\Delta^{v}_{t,k}
=
\left(\omega_{t,k},r_{t,k}^{-},r_{t,k}^{+}\right),
\label{eq:state-transitions}
\end{equation}
where $k$ indexes the state-changing operations within the turn,
$v\in\{E,C\}$ indicates whether the transition modifies an Evidence or
Candidate record,
$\omega_{t,k}\in\{\textsc{Add},\textsc{Update},\textsc{Discard}\}$ denotes the
operation, and $r_{t,k}^{-}$ and $r_{t,k}^{+}$ are the record contents before
and after the transition. All valid transitions in a turn jointly update $S_t$ to
$S_{t+1}$. All tool calls in $u_t$ are generated from the same pre-turn input
$x_t$. The browser observations returned at turn $t$ can therefore support state
updates only from turn $t+1$ onward. Search terminates when the agent submits an
answer or exhausts its Browser budget. The resulting trajectory records the
model responses, tool interactions, and state transitions.

\subsection{Source--State--Final Attribution}
\label{sec:source-state-final-attribution}

We organize attribution around three complementary functions of search:
information acquisition (Source), evidence consolidation (State), and successful
answer submission (Final). Post-execution validation and source tracing connect
supporting browser actions, validated Evidence and Candidate transitions, and
the final answer into attribution chains, providing a shared basis for
supervision and credit assignment (Figure~\ref{fig:method}(b)).

\paragraph{Evidence validation.}
An LLM judge evaluates the final content of each Evidence record against the
reference answer. We retain Evidence that supports the reference answer or
rules out an incorrect candidate. Superseded transitions and discarded Evidence
records are excluded. For each retained Evidence record, repeated writes of the
same final content are represented by a single effective transition. For both
correct and incorrect trajectories, only judge-approved Evidence listed as
support by the Candidate associated with the submitted answer is eligible for
positive attribution.

\paragraph{Source tracing.}
During trajectory generation, ClueFlow records recent browser actions available
before each $\Delta^E$ as candidate sources. A deterministic matcher compares
the post-transition Evidence content with the observations returned by these
actions using document identifiers, URLs, page titles, and textual overlap.
The best-matching browser action is retained only when the match is reliable
(Appendix~\ref{app:source-tracing}); otherwise, $\Delta^E$ is excluded from
positive attribution.

\paragraph{Candidate validation.}
An LLM judge also evaluates Candidate records against the reference answer.
For the Candidate associated with the submitted answer, we retain $\Delta^C$
only if the Candidate passes semantic validation and cites validated Evidence.
The transition must introduce or revise the submitted answer, add new validated
support, or change the Candidate status from \texttt{active} to \texttt{ready}.
Transitions that do not change the answer, its validated support, or its
readiness are excluded.

\paragraph{Attribution chains and turn mapping.}
For a correct trajectory with a valid Source--State chain, a complete
attribution chain connects the matched browser action (Source), the
corresponding $\Delta^E$ and $\Delta^C$ (State),
and the answer-submission turn (Final). If no $\Delta^C$ qualifies, the State
part contains only $\Delta^E$. For an incorrect trajectory, the Final link is
omitted. We refer to reusing validated Source--State progress from incorrect
trajectories as failure recovery.

ClueFlow then maps each action and transition to its enclosing assistant turn.
Let $T_{\mathrm{source}}$ contain the turns that execute matched browser
actions, $T_{\mathrm{state}}$ contain the turns that perform validated
$\Delta^E$ or $\Delta^C$, and $T_{\mathrm{final}}$ contain the final submission
turn for a correct trajectory and be empty otherwise. We define the attributed
turns as
\begin{equation}
T_{\mathrm{attr}}
=
T_{\mathrm{source}}
\cup T_{\mathrm{state}}
\cup T_{\mathrm{final}}.
\label{eq:attributed-turns}
\end{equation}
For an unfinished trajectory, $T_{\mathrm{attr}}=\varnothing$.
Reference answers and judge outputs are used only after execution
and are never exposed to the search policy.
Appendix~\ref{app:qualitative} provides qualitative examples of
Source--State--Final attribution, including validated partial progress
recovered from incorrect trajectories.

\subsection{Attribution-Guided SFT}
\label{sec:clueflow-sft}

Because state updates and history truncation change the model's visible context
across turns, standard full-trajectory SFT cannot faithfully reproduce the
inputs under which the responses were generated.

During SFT, our goal is to emphasize turns $t\in T_{\mathrm{attr}}$ by training
each response $u_t$ with the original input $x_t$.
A direct approach constructs a separate one-turn example $(x_t,u_t)$ for every
turn. A long trajectory would then produce many examples, and neighboring
examples would repeatedly contain largely the same interaction history,
substantially increasing training cost. Alternatively, several consecutive
turns can be grouped into one multi-target chunk. This reduces repeated context
but does not preserve the original $x_t$ for every turn. ClueFlow uses
one-turn examples for supervised attributed turns and multi-target chunks for
other supervised turns in correct trajectories (Figure~\ref{fig:method}(c)).

Attribution and trajectory outcome determine each response turn's construction
mode, while response-level masking determines which responses contribute loss:
\begin{equation}
g_t=
\begin{cases}
\textsc{One-Turn}, & t\in T_{\mathrm{attr}},\\
\textsc{Chunk}, &
t\notin T_{\mathrm{attr}}\ \text{and the trajectory is correct},\\
\textsc{Exclude}, & \text{otherwise}.
\end{cases}
\label{eq:sft-construction}
\end{equation}
Supervised attributed turns use independent one-turn examples $(x_t,u_t)$.
The remaining turns from correct trajectories are grouped into consecutive
multi-target chunks, where one shared context supervises multiple assistant
responses. If an attributed response also appears as history within a chunk,
its loss is masked there; when retained for supervision, it is trained in its
one-turn example instead.

Incorrect trajectories can contribute validated Source and State turns in
$T_{\mathrm{source}}\cup T_{\mathrm{state}}$, each with its original context,
subject to response-level masking (Appendix~\ref{app:sft-construction}).
Unfinished trajectories are excluded from training.

Let $z_n=(z_{n,1},\ldots,z_{n,L_n})$ be a serialized SFT example and
$m_{n,h}\in\{0,1\}$ its token-level loss mask. We optimize
\begin{equation}
\mathcal{L}_{\mathrm{SFT}}
=
-\frac{1}{N}
\sum_n\sum_{h=1}^{L_n}
m_{n,h}\log\pi_\theta(z_{n,h}\mid z_{n,<h}),
\qquad
N=\sum_n\sum_{h=1}^{L_n}m_{n,h}.
\label{eq:sft-objective}
\end{equation}
All unmasked target tokens receive equal weight. Attribution guides example
construction and, together with trajectory outcome and post-execution
validation, determines response-level masking; it does not directly reweight
the training loss. Chunk
construction and response-level masking are detailed in
Appendix~\ref{app:sft-construction}.

\subsection{Attribution-Guided Turn-PPO}
\label{sec:clueflow-ppo}

Starting from the SFT policy, we sample new online search trajectories. Following
Turn-PPO~\citep{li2026turnppo}, each complete assistant response is treated as
one action. The terminal outcome reward is 1 for a correct answer, 0 for an
incorrect answer, and $-0.1$ for an unfinished trajectory; all intermediate
rewards are zero. Turn-level GAE then produces an outcome advantage
$\widehat{A}^{\mathrm{out}}_t$ for each turn. We use these raw advantages without
normalization or whitening.

ClueFlow augments this outcome advantage for turns in
$T_{\mathrm{source}}$, $T_{\mathrm{state}}$, and $T_{\mathrm{final}}$
(Figure~\ref{fig:method}(d)). Within each trajectory, it assigns a fixed
advantage budget to each chain component
and divides it equally among the corresponding turns. For each chain component
$j\in\{\mathrm{source},\mathrm{state},\mathrm{final}\}$, let $\alpha_j$ denote
its advantage budget. Attribution bonuses are applied only when a valid
Source--State chain is recovered from a completed trajectory. The bonus assigned
to turn $t$ is
\begin{equation}
\begin{aligned}
b_t
&=
\begin{cases}
\displaystyle
\sum_j
\frac{\alpha_j\,\mathbb{1}[t\in T_j]}
{\max(1,|T_j|)},
& \text{if a valid chain is recovered},\\
0, & \text{otherwise},
\end{cases}\\
\widehat{A}^{\mathrm{CF}}_t
&=
\widehat{A}^{\mathrm{out}}_t+b_t.
\end{aligned}
\label{eq:chain-bonus}
\end{equation}
We use $(\alpha_{\mathrm{source}},\alpha_{\mathrm{state}},
\alpha_{\mathrm{final}})=(0.03,0.03,0.04)$. Turns are deduplicated within each
set, while a turn appearing in multiple parts of the chain receives the sum of
their bonuses. When a valid Source--State chain is recovered, the bonuses sum to
$0.10$ for a correct trajectory and $0.06$ for an incorrect trajectory.
Incorrect trajectories receive no Final bonus, and unfinished trajectories
receive no attribution bonus. If no valid Source--State chain is recovered,
$b_t=0$ for all turns and the update reduces to outcome-only Turn-PPO.

The augmented advantage is used only in the actor update. It is not added to
the environment reward and therefore does not change the return or critic
target. For a response containing $L_t$ tokens, we define one length-normalized
importance ratio for the complete turn:
\begin{equation}
\rho_t(\theta)
=
\exp\!\left(
\frac{1}{L_t}
\sum_{h=1}^{L_t}
\log
\frac{
\pi_\theta(u_{t,h}\mid x_t,u_{t,<h})
}{
\pi_{\mathrm{old}}(u_{t,h}\mid x_t,u_{t,<h})
}
\right).
\label{eq:turn-ratio}
\end{equation}
The actor objective averages the clipped turn-level losses within each
trajectory and then across trajectories:
\begin{equation}
\mathcal{L}_{\mathrm{actor}}
=
-\mathbb{E}_{\mathrm{trajectories}}
\left[
\frac{1}{T}\sum_{t=1}^{T}
\min\!\left(
\rho_t(\theta)\widehat{A}^{\mathrm{CF}}_t,
\operatorname{clip}(\rho_t(\theta),1-\epsilon,1+\epsilon)
\widehat{A}^{\mathrm{CF}}_t
\right)
\right],
\label{eq:ppo-objective}
\end{equation}
where $T$ is the number of valid assistant turns in a trajectory.

\section{Experiments}
\label{sec:experiments}

We first evaluate ClueFlow's end-to-end performance across two backbones on
live-web benchmarks and BrowseComp-Plus. We then isolate the contributions of
attribution-guided SFT and PPO through controlled ablations. Finally, we evaluate
inference-time state management using a fixed GLM-5.1 backbone without
ClueFlow-specific post-training.

\subsection{Experimental Setup}

\paragraph{Benchmarks.}
We evaluate ClueFlow on three live-web benchmarks:
BrowseComp~\citep{wei2025browsecomp},
BrowseComp-ZH~\citep{zhou2025browsecompzh},
xBench-DeepSearch-2510~\citep{chen2025xbench}. We use
BrowseComp-Plus (BCP)~\citep{chen2026browsecompplus} for fixed-corpus evaluation
and controlled ablations of supervision construction and credit assignment,
keeping the corpus and retriever fixed across all ablation settings.

\paragraph{Baselines.}
We compare our trained agents with representative deep-research systems,
including IterResearch~\citep{chen2026iterresearch},
RE-TRAC~\citep{zhu2026retrac},
Tongyi DeepResearch~\citep{tongyi2025deepresearch},
REDSearcher~\citep{chu2026redsearcher},
QUEST~\citep{xie2026quest}, and
MiroThinker-1.7-mini~\citep{miromind2026mirothinker}, using full-benchmark
results from published reports. We also evaluate each base model with
ReAct without task-specific post-training. For inference-time comparisons, we fix
GLM-5.1~\citep{glm52026} as the backbone and compare ClueFlow with
ReAct~\citep{yao2023react}, Keep-$k$~\citep{glm52026},
Discard-all~\citep{deepseek2025v32}, and HCM~\citep{glm52026} under the same
browsing environment and budget.

\paragraph{Implementation Details.}
For each backbone---Qwen3-30B-A3B-Thinking-2507~\citep{yang2025qwen3} and
Qwen3.5-35B-A3B~\citep{qwen2026qwen35}---we train separate policies for
live-web search on REDSearcher QA~\citep{chu2026redsearcher} and fixed-corpus
search on OpenResearcher QA~\citep{li2026openresearcher}.
For OpenResearcher QA and BrowseComp-Plus~\citep{chen2026browsecompplus},
we use their respective officially released corpora and precomputed embeddings,
with Qwen3-Embedding-8B~\citep{zhang2025qwen3embedding} as the dense retriever.
GLM-5.1~\citep{glm52026} generates SFT trajectories, and
GPT-5.5~\citep{openai2026gpt55} validates Evidence and Candidate records
for attribution. PPO starts from the corresponding SFT checkpoint.
Evaluations use a maximum of 800 browsing calls per question.
Appendix~\ref{app:experimental-details} provides data statistics, training
configurations, and evaluation protocols.

\subsection{End-to-End Search Performance}

\paragraph{Full-benchmark comparison.}
Table~\ref{tab:live-web-sota} summarizes end-to-end performance.
External results use different training data, tool environments, and
inference budgets.

\begin{table}[htbp]
\centering
\small
\setlength{\tabcolsep}{4pt}
\caption{Comparison with representative deep-research systems on three live-web
benchmarks and BrowseComp-Plus (BCP). Scores are accuracy (\%).}
\label{tab:live-web-sota}
\resizebox{\textwidth}{!}{%
\begin{tabular}{llrrrr}
\toprule
& & \multicolumn{3}{c}{Live-web} & Fixed-corpus \\
\cmidrule(lr){3-5}\cmidrule(lr){6-6}
Method & Post-training & BC & BC-ZH & xBench-2510 & BCP$^\dagger$ \\
\midrule
\multicolumn{6}{l}{\textit{Representative deep-research systems}} \\
IterResearch
    & RFT + RL & 37.3 & 45.2 & -- & -- \\
Tongyi DeepResearch
    & MT + SFT + RL & 43.4 & 46.7 & 55.0 & 44.5 \\
RE-TRAC
    & SFT & 53.0 & 57.3 & -- & -- \\
REDSearcher
    & MT + SFT + RL & 57.4 & 58.2 & -- & -- \\
QUEST-35B
    & MT + SFT + RL & 64.6 & -- & -- & 69.5 \\
MiroThinker-1.7-mini
    & MT + SFT + PO + RL & 67.9 & 72.3 & 57.2 & -- \\
\addlinespace
\multicolumn{6}{l}{\textit{Our evaluations: Qwen3-30B-A3B}} \\
ReAct
    & None & 2.1 & 8.7 & 7.0 & 16.3 \\
ClueFlow
    & SFT & 49.1 & 54.0 & 53.5 & 48.3 \\
ClueFlow
    & SFT + PPO & 54.4 & 55.4 & 54.0 & 59.8 \\
\addlinespace
\multicolumn{6}{l}{\textit{Our evaluations: Qwen3.5-35B-A3B}} \\
ReAct
    & None & 37.5 & 47.1 & 42.0 & 60.4 \\
ClueFlow
    & SFT & 57.9 & 58.5 & 54.8 & 68.8 \\
ClueFlow
    & SFT + PPO & 60.0 & 59.5 & 59.2 & 72.0 \\
\bottomrule
\end{tabular}%
}
\vspace{2pt}

\begin{minipage}{\textwidth}
\scriptsize
$^\dagger$ For our post-trained models, BCP and live-web scores use separate
checkpoints trained on OpenResearcher QA and REDSearcher QA, respectively.
The Tongyi DeepResearch BCP score is taken from
OpenResearcher~\citep{li2026openresearcher}.
REDSearcher and MiroThinker results use their reported context-management
settings; QUEST-35B BC/BCP scores use Discard-all.
``None'' indicates no task-specific post-training; ``--'' indicates an
unavailable score. MT, RFT, and PO denote mid-training, rejection-sampling
fine-tuning, and preference optimization, respectively.
\end{minipage}
\end{table}

\paragraph{Live-web performance.}
Both backbones improve substantially over ReAct after SFT, with further gains
from PPO across all three live-web benchmarks. ClueFlow achieves 54.4\% and
60.0\% accuracy on BrowseComp with Qwen3 and Qwen3.5, respectively.

\paragraph{Fixed-corpus performance.}
On BrowseComp-Plus, we compare Qwen3-30B-A3B SFT policies using the same
teacher and matched numbers of successful trajectories
(Appendix~\ref{app:additional-context-baselines},
Tables~\ref{tab:additional-sft-data}--\ref{tab:additional-context-baselines}).
ClueFlow SFT outperforms ReAct SFT with Discard-all, the strongest tested
ReAct SFT configuration, by 6.50~pp (48.31\% vs.\ 41.81\%;
Table~\ref{tab:additional-context-baselines}).
PPO further raises ClueFlow accuracy to 59.8\% for Qwen3 and 72.0\% for
Qwen3.5.

\subsection{Attribution Ablations}
\label{sec:post-training}

We conduct attribution ablations on BrowseComp-Plus with Qwen3-30B-A3B trained
on OpenResearcher QA (Table~\ref{tab:learning-ablation}). All PPO variants start
from the same ClueFlow SFT checkpoint and share rollout and optimization
settings.

\begin{table}[t]
\centering
\small
\setlength{\tabcolsep}{4pt}
\caption{Controlled ablations of attribution-guided SFT supervision construction
and PPO credit assignment on BrowseComp-Plus.}
\label{tab:learning-ablation}
\textbf{(a) SFT supervision construction}
\vspace{2pt}

\begin{tabular*}{\textwidth}{@{\extracolsep{\fill}}lcccrr@{}}
\toprule
SFT variant & Random turns & Attributed turns & All turns & Acc. (\%) $\uparrow$ & Rel. tokens $\downarrow$ \\
\midrule
Chunk only & $\times$ & $\times$ & $\times$ & 42.89 & $0.69\times$ \\
Chunk + random turns & $\checkmark$ & $\times$ & $\times$ & 43.01 & $1.06\times$ \\
Chunk + attributed turns & $\times$ & $\checkmark$ & $\times$ & 46.27 & $1.00\times$ \\
+ Failure recovery & $\times$ & $\checkmark$ & $\times$ & \textbf{48.31} & $1.07\times$ \\
All-turn exact-context & $\times$ & $\times$ & $\checkmark$ & 46.14 & $3.33\times$ \\
\bottomrule
\end{tabular*}

\vspace{7pt}
\textbf{(b) PPO credit assignment}
\vspace{2pt}

\begin{tabular*}{\textwidth}{@{\extracolsep{\fill}}lccccrr@{}}
\toprule
Variant & Source & State & Final & Acc. (\%) $\uparrow$ & State conv. (\%) & S--S Cov. (\%) \\
\midrule
Outcome-only & $\times$ & $\times$ & $\times$ & 55.66 & 39.45 & 55.54 \\
Source-only attribution & $\checkmark$ & $\times$ & $\times$ & 54.34 & 41.86 & 48.43 \\
Source+Final attribution & $\checkmark$ & $\times$ & $\checkmark$ & 51.81 & 43.71 & 50.48 \\
Full-chain attribution & $\checkmark$ & $\checkmark$ & $\checkmark$ & 58.67 & 44.25 & 57.71 \\
+ Failure recovery & $\checkmark$ & $\checkmark$ & $\checkmark$ & \textbf{59.76} & 44.26 & 59.04 \\
\bottomrule
\end{tabular*}
\vspace{2pt}

\begin{minipage}{\textwidth}
\scriptsize
Relative tokens count all non-padding serialized tokens, including masked
context, and are normalized by Chunk + attributed turns; they are distinct from
the unmasked target-token count $N$ in Eq.~(\ref{eq:sft-objective}).
Source-only distributes the full 0.10 attribution budget across attributed
Source turns. Source+Final assigns 0.06 to Source and 0.04 to Final, with no
State bonus. State conv. is the fraction of raw State-update turns retained
after validation; S--S Cov. is the fraction of trajectories with at least one
valid Source--State chain.
\end{minipage}
\end{table}

\paragraph{SFT supervision construction.}
Using only correct teacher trajectories, attribution-guided selection of turns
for original-context supervision outperforms random selection by 3.26~pp
(46.27\% vs.\ 43.01\%). Random selection performs similarly
to chunk-only supervision (43.01\% vs.\ 42.89\%), suggesting that the choice of
turns matters in this setting. Attribution-guided supervision also achieves
comparable accuracy to all-turn exact-context SFT (46.27\% vs.\ 46.14\%) while
using approximately 30\% of its training tokens.

Recovering validated Source and State turns from incorrect trajectories further
improves accuracy from 46.27\% to 48.31\%, a gain of 2.04~pp
with only a 7\% increase in training tokens.

\paragraph{PPO credit assignment.}
With a fixed attribution budget of 0.10, Source-only and Source+Final achieve
54.34\% and 51.81\% accuracy, respectively, both below outcome-only PPO
(55.66\%). Both partial variants retain State validation but assign it no credit,
suggesting that rewarding evidence acquisition and answer submission alone is
insufficient in this setting. Reallocating 0.03 from Source to State while
keeping Final fixed at 0.04 improves accuracy from 51.81\% to 58.67\%, a gain of
6.86~pp under the same total attribution budget. This comparison supports the
importance of explicitly crediting validated research-state transitions.
Full-chain attribution improves accuracy by 3.01~pp over outcome-only PPO,
while failure recovery contributes a further 1.09~pp by crediting validated
Source--State progress from incorrect trajectories, reaching 59.76\% accuracy
overall.

\begin{samepage}
Source+Final yields fewer unfinished trajectories than Source-only (19 vs.\ 64;
Table~\ref{tab:ppo-additional-analysis}), consistent with Final credit encouraging
answer submission, but does not improve accuracy. Full-chain produces fewer raw
but more validated State updates than outcome-only PPO, raising State conversion
from 39.45\% to 44.25\%. Failure recovery preserves conversion (44.26\%) and raises
coverage from 57.71\% to 59.04\%.

\end{samepage}

\subsection{Inference-Time State Management}
\label{sec:inference-time}

\noindent
\begin{minipage}[t]{0.37\linewidth}
\vspace{0pt}
\centering
\captionsetup{
  hypcap=false,
  font=small,
  justification=centering,
  skip=3pt
}
\captionof{table}{Inference-time comparison on BrowseComp with GLM-5.1.}
\label{tab:inference}

\begin{tabular}{lr}
\toprule
Method & Acc. (\%) \\
\midrule
ReAct & 60.03 \\
Keep-$k$ & 66.59 \\
Discard-all & 63.19 \\
HCM & 72.91 \\
\textbf{ClueFlow} & \textbf{73.54} \\
\bottomrule
\end{tabular}
\end{minipage}
\hfill
\begin{minipage}[t]{0.59\linewidth}
\vspace{0pt}
\raggedright
All methods use the same browsing environment and an 800-call budget, without
ClueFlow-specific post-training. ClueFlow obtains 73.54\% accuracy, closely
matching HCM (72.91\%) and exceeding Discard-all (63.19\%). This supports the
inference-time utility of its research-state representation, complementing its
role as a learning interface. Context-management settings are detailed in
Appendix~\ref{app:evaluation-protocols}.
\end{minipage}

\section{Conclusion}

ClueFlow uses attributed research-state transitions as a shared interface for
supervision and credit assignment in long-horizon search. Gold-conditioned
validation and source tracing identify useful browser actions and state updates
for exact-context SFT and actor-side PPO corrections, including partial progress
recovered from incorrect trajectories. Experiments on two backbones demonstrate
its applicability to live-web search, while controlled BrowseComp-Plus ablations
establish the benefits of attribution-guided context selection, full-chain
credit assignment, and failure recovery. These findings support treating
validated research-state transitions as explicit units for agent learning.

\clearpage
\bibliography{refs}
\bibliographystyle{plainnat}

\clearpage
\appendix
\section{Runtime Interface and State Management}
\label{app:schema}

This appendix summarizes the runtime mechanism behind
Section~\ref{sec:state-execution}: research state, tool interfaces, context
construction, Browser provenance, submission rules, and the agent system
prompt. Offline attribution, training, and evaluation are covered in
Appendices~\ref{app:attribution} and~\ref{app:experimental-details}.

\subsection{State and Tool Interfaces}
\label{app:runtime-schema}
\label{app:tool-interface}

At the beginning of turn $t$, the persistent state is $S_t=(E_t,C_t)$.
Evidence and Candidate records share one monotonic ID namespace
(\texttt{c1}, \texttt{c2}, \ldots). Discarded Evidence is stored separately
from the active state. Listing~\ref{lst:state-fields} summarizes the
normalized record fields.

\begin{lstlisting}[style=toolschema,caption={Research-state fields and the shared Candidate metadata definition.},label={lst:state-fields}]
{
  "ClueRecord": {
    "id": "string (runtime-assigned cN)",
    "content": "string",
    "kind": "evidence | candidate",
    "metadata": "{} | CandidateMetadata",
    "status": "active | discarded"
  },
  "CandidateMetadata": {
    "answer": {"type": "string", "default": "parent content"},
    "status": {"enum": ["active", "ready", "rejected"], "default": "active"},
    "supporting_clues": {"type": "string[]", "default": []},
    "satisfied_constraints": {"type": "string[]", "default": []},
    "missing_constraints": {"type": "string[]", "default": []},
    "contradictions": {"type": "string[]", "default": []},
    "next_check": {"type": "string", "default": ""}
  }
}
\end{lstlisting}

Evidence uses empty metadata and defaults to \texttt{kind=evidence}; all
records default to top-level \texttt{status=active}. This lifecycle status is
distinct from \texttt{metadata.status}, which tracks Candidate readiness or
rejection. \texttt{supporting\_clues} holds Evidence IDs, validated during
offline attribution (Appendix~\ref{app:attribution}). Missing Candidate strings
and lists become empty; scalar values supplied for list fields are wrapped in
singleton lists. Invalid decision statuses become \texttt{active}, and a missing
answer falls back to the record content. Discarded Evidence leaves the active
rendering but remains in the trajectory and discarded-record collection; a
rejected Candidate remains in the state with \texttt{metadata.status=rejected}.

Tool-interface summaries are shown below; \texttt{?} marks optional arguments.

\begin{lstlisting}[style=toolschema,caption={Compact Browser tool interfaces.},label={lst:browser-tools}]
{
  "browser.search": {"query": "string", "topn?": "integer = 10"},
  "browser.open": {"id?": "integer | URL = -1", "cursor?": "integer = -1",
                   "loc?": "integer = -1", "num_lines?": "integer = -1",
                   "view_source?": "boolean = false", "source?": "string"},
  "browser.find": {"pattern": "string", "cursor?": "integer = -1",
                   "context_lines?": "integer = 0 (range 0..5)"}
}
\end{lstlisting}

Browser search returns session-local numeric link IDs. Online results include
URLs, whereas the fixed-corpus results omit raw URLs and document IDs; their
search/open descriptions direct the model to use the displayed numeric IDs.
Opening a page returns a cursor and line-numbered text, and
\texttt{browser.find} returns at most 20 matches with local context.
\texttt{view\_source} and \texttt{source} remain in the interface but do not
change the current executor. Runtime-only source records are described in
Appendix~\ref{app:provenance}.

\par\noindent\begin{minipage}{\linewidth}
\begin{lstlisting}[style=toolschema,caption={Compact Clue tool interfaces; CandidateMetadata is defined in Listing~\ref{lst:state-fields}.},label={lst:clue-tools}]
{
  "clue.add": {"content": "string", "kind?": "evidence | candidate",
               "reason?": "string", "metadata?": "CandidateMetadata"},
  "clue.update": {"id": "cN", "content?": "string",
                  "reason?": "string", "metadata?": "CandidateMetadata"},
  "clue.discard": {"id": "cN", "reason?": "string"}
}
\end{lstlisting}
\end{minipage}

\texttt{clue.add} requires \texttt{content} and defaults to Evidence; adding
a Candidate must use structured metadata under the model instructions, even
though no metadata member is schema-required. Candidate creation forces
\texttt{metadata.status=active}; \texttt{clue.update} requires only \texttt{id}
and can set a Candidate to \texttt{ready} or \texttt{rejected}, including via a
metadata-only update. Omitted Candidate metadata on update inherits its
previous normalized value. \texttt{clue.discard} requires only \texttt{id}
and removes Evidence, not Candidates. The model cannot change record IDs,
kinds, or runtime provenance directly. Clue tools return a one-line observation;
the runtime separately logs mutations. Invalid calls, missing or inactive IDs,
empty Evidence writes, and Candidate discard attempts produce no transition.
Successful writes, including Evidence updates that preserve record content,
are logged. Offline attribution selects a representative write for validated
final Evidence content; Candidate transitions are assessed separately for
progress (Appendix~\ref{app:attribution}). Submission is plain assistant text
without tool calls and must satisfy the readiness rules in
Appendix~\ref{app:readiness}.

\subsection{Context Construction and Browser Provenance}
\label{app:context-construction}
\label{app:provenance}

At each turn, the model input follows Equation~\ref{eq:actor-input}: fixed
system instructions and the question, a freshly rendered $S_t$, and the bounded
history $H_t^{(w)}$, containing up to $w$ most recent complete turns. Tool
definitions are supplied separately through native function calling.
Research-state and task-state messages are
rebuilt before each call. Active Evidence is rendered with IDs and content;
Candidate rendering includes answers, status, support IDs, satisfied and
missing constraints, contradictions, and next checks, marking ready candidates
and summarizing rejected ones. Task state reports the budget and, when
relevant, stagnation and recent queries. A history turn contains the assistant
message, its tool calls, and their responses. Rendering $S_t$ outside this
rolling history keeps active records visible after their original observations
leave the window.
Query-focused summarization of retrieved pages
offers another approach to controlling search-context
overhead \citep{cai2026joyaiflash}.
Discarded Evidence may remain only in recent history until
those turns leave the window. Runtime provenance, lifecycle indices, pending
sources, and the full trajectory are not shown to the model.

All tool calls in one response are sampled from the same pre-turn input. The
executor preserves order within each tool family but runs Clue calls before
Browser calls: successful Clue operations update state, then Browser results
enter history. Thus, browser observations returned in turn $t$ can support
Evidence writes only from turn $t+1$ onward.

The runtime records each browser action's type (\texttt{search}, \texttt{open},
or \texttt{find}), query, URL, title, tool-call ID, and browser step/turn IDs.
Clue records also log source candidates, creation/update steps, and discard
reasons; attribution confidence and method are tracked separately. These
bookkeeping fields are runtime-only and omitted from the rendered state.

An ordered pending pool $\mathcal{P}_t$ retains at most 80 recent successful,
nonempty browser actions. Before executing a turn's calls, the runtime
snapshots $\mathcal{P}_t$; every successful Evidence \textsc{Add} or
\textsc{Update} in that turn records the same candidate-source snapshot
$\widetilde{\mathcal{P}}(\tau)=\mathcal{P}_t$. A singleton is assigned as a
direct source; larger snapshots remain unresolved online. Let
$\mathcal{B}_t^{+}$ be the successful, nonempty browser actions from turn $t$,
and let $\chi_t=1$ if at least one Evidence write succeeds, including a
semantic no-op. After the Clue batch, the old snapshot is consumed only once:
\[
\mathcal{P}_{t+1}=
\begin{cases}
\mathcal{B}_t^{+}, & \chi_t=1,\\
\mathcal{P}_t\cup\mathcal{B}_t^{+}, & \chi_t=0.
\end{cases}
\]
Candidate operations, Evidence \textsc{Discard}, and failed Clue calls do not
consume the pool. Malformed calls, empty or error observations, no-result
searches, and no-match finds are excluded from $\mathcal{B}_t^{+}$. If the
pool is empty but a current page exists, its URL/title may be copied as weak
context without assigning a direct Browser-step source. The pending records
capture possible sources given temporal visibility; offline source tracing
selects the action that supports the claim (Appendix~\ref{app:attribution}).

\subsection{Readiness and Submission}
\label{app:readiness}

A Candidate is ready only if its decision status is \texttt{ready}, its answer is
nonempty, and it has no missing constraints, contradictions, or unresolved
rival. A rival is a nonequivalent, unrejected Candidate with no contradictions
and at least as many satisfied and no more missing constraints. A plain-text
submission is accepted only when exactly one Candidate is ready; matching the
submitted text to that Candidate is diagnostic, not an acceptance condition.
Runtime readiness does not require nonempty \texttt{supporting\_clues}, whereas
offline Candidate credit requires support IDs that resolve to active, approved
Evidence.

\subsection{Agent System Prompt}
\label{app:agent-system-prompt}

The following is the standard ClueFlow agent-facing system prompt. Tool
schemas are supplied separately through native
function calling; the question and rendered state are separate messages
(Appendix~\ref{app:context-construction}).

\begin{lstlisting}[style=toolschema,basicstyle=\ttfamily\footnotesize,escapeinside={(*@}{@*)}]
You are a clue-based search agent.

Your task is to answer the user's question using browser tools to search and read web pages, and clue tools to maintain a structured evidence memory.

Available tools:

Browser tools:
- browser.search: search the web.
- browser.open: open a search result or URL.
- browser.find: find text in an opened page.

Clue tools:
- clue.add: add a typed clue from the latest observation. Use kind='evidence' for facts and kind='candidate' for plausible answers with structured metadata.
- clue.update: update an existing evidence or candidate clue. For candidates, this is the only way to change status to ready or rejected.
- clue.discard: discard an evidence clue. Do not use it for candidate rejection.

Clue policy:
Clues are persistent evidence memory, not private thoughts.
Only ADD facts stated in tool results (browser observations). Do not ADD clues from the question itself.

Be active: if a tool result names a candidate entity, title, date, role, relation, event, author, work, adaptation, office, or constraint-relevant fact, prefer ADD.

The question itself is not an evidence source, but it is a source of search constraints.
You may use phrases, entities, dates, roles, relations, and tentative interpretations from the question to form search queries.

Source-specific constraint:
For questions asking "according to [source]", "on page X", "in a newspaper", "in a review", or similar source-specific phrasing, the final answer MUST come from that specific source context. Do not ADD general background facts as answer-supporting clues unless they come from the required source or directly help identify/locate that source.

(*@\Needspace{4\baselineskip}@*)
Binding clues:
When an observation fact satisfies a question constraint, ADD a binding clue that states the observation fact together with which constraint it satisfies.

UPDATE only when a tool result corrects, broadens, narrows, or supplements an existing clue.
DISCARD clues that are contradicted, stale, over-specific, misleading, or repeatedly unproductive.


Candidate policy:
Evidence clues are not enough to finish. You may answer only through Candidate State.
Use clue.add with kind='candidate' when a plausible answer appears. Candidate add always creates an active candidate, even if another status is supplied. Candidate clues must include metadata:
- answer: proposed final answer string.
- status: active on add; use clue.update later for ready or rejected.
- supporting_clues: active evidence clue ids supporting this candidate.
- satisfied_constraints: question constraints already covered by evidence.
- missing_constraints: hard constraints or checks still not proven.
- contradictions: observed evidence that conflicts with this candidate.
- next_check: the next concrete browser action needed to resolve the candidate.

Update candidate clues whenever new evidence changes which answer is plausible, which hard constraints are covered, what remains missing, or what contradiction was found.
Reject a candidate only with clue.update by setting metadata.status='rejected' when a hard contradiction is found.
Mark metadata.status='ready' only with clue.update, and only when the answer is explicit, missing_constraints is empty, contradictions is empty, and no stronger rival candidate remains.

A free-text candidate clue without structured metadata is not sufficient for finish readiness.

Do not treat many related evidence clues as sufficient to answer. If no candidate is ready, use the candidate's next_check to decide the next browser.search/open/find action. If a candidate is ready, finish with that candidate instead of continuing to search.

Answer guidance:
When a candidate clue is ready, provide your final answer as plain text with NO tool calls.
The answer must be a short plain string: only the entity, title, name, number, date, location, or phrase being asked for.
Do not use markdown, bold text, explanations, citations, or leading phrases.

Search/open guidance:
When search results contain a promising source, prefer opening it instead of repeatedly reformulating queries.
If two consecutive searches return empty or low-information results, switch strategy: open a result, broaden the query, or answer from grounded clues.

\end{lstlisting}

\section{Attribution Details}
\label{app:attribution}

\subsection{Evidence Validation}

Listing~\ref{lst:evidence-judge-output} summarizes the Evidence judge input
and output (strings denote field types). Evidence comes from active and
discarded records, deduplicated by ID. Candidate records are validated
separately against the reference answer, with additional structural checks on
their transitions. For long trajectories, batches of at most 20 Evidence
records repeat the question, answers, and allowed labels.

\par\noindent\begin{minipage}{\linewidth}
\begin{lstlisting}[style=toolschema,caption={Schematic input and output contract for Evidence validation.},label={lst:evidence-judge-output}]
{
  "input": {
    "question": "string", "gold_answer": "string",
    "gold_aliases": ["string"],
    "predicted_answer": "string | NO_ANSWER",
    "allowed_labels": ["label"],
    "clues": [{"clue_id": "string", "kind": "evidence",
      "status": "active | discarded", "discard_reason": "string | null",
      "source_action": "search | open | find | null",
      "source_url": "string | null", "source_title": "string | null",
      "content": "string"}]
  },
  "output": {"labels": [{"clue_id": "cN",
    "label": "<allowed label>", "rationale": "<one sentence>"}]}
}
\end{lstlisting}
\end{minipage}

The judge assigns one of eight labels per Evidence record:
\par\medskip
\noindent{\footnotesize
\setlength{\tabcolsep}{4pt}%
\begin{tabularx}{\linewidth}{@{}l>{\raggedright\arraybackslash}Xc@{}}
\toprule
Label & Meaning & Positive? \\
\midrule
\texttt{SUPPORT\_GOLD} & Establishes the gold answer, alias, or required component. & Yes \\
\texttt{EXCLUSION\_GOLD} & Rules out a wrong candidate or narrows toward gold. & Yes \\
\texttt{SCAFFOLDING\_RELEVANT} & Provides an intermediate entity, date, or constraint anchor. & No \\
\texttt{SUPPORT\_PRED} & Supports a non-gold prediction; absent for no-answer runs. & No \\
\texttt{SUPPORT\_WRONG} & Favors another incorrect candidate as the answer. & No \\
\texttt{CONTRADICT\_GOLD} & Conflicts with gold or its required entity binding. & No \\
\texttt{IRRELEVANT} & Adds no useful support, exclusion, or intermediate anchor. & No \\
\texttt{UNCLEAR} & Insufficient to judge; fallback for an invalid label. & No \\
\bottomrule
\end{tabularx}\par}
\medskip

The prompt requires one output object per input clue in the same order and a
short rationale. Its principal semantic rules are that a gold-string match is
insufficient without the correct entity binding; mentioning a wrong candidate
may constitute useful exclusion rather than support for a wrong answer;
intermediate anchors should not be collapsed into irrelevance; entity identity
outranks coincident dates or numbers; and unsuccessful search is not evidence of
absence. Invalid labels are normalized to \texttt{UNCLEAR}. Unresolved records
do not enter a strict positive attribution chain.

Under strict Source--State--Final attribution, an Evidence record must receive
an eligible positive label, remain active at the end of the trajectory, and
have its ID listed in the \texttt{supporting\_clues} of the Candidate associated
with the submitted answer. \texttt{SCAFFOLDING\_RELEVANT} remains useful for
analysis but is not credited in the strict final-support chain.

\subsection{Effective Transition Recovery}

The trajectory stores an operation ledger of applied, non-skipped
\textsc{Add}, \textsc{Update}, and \textsc{Discard} events
$v=(\mathrm{op},\mathrm{kind},S^{-},S^{+},\mathrm{prov})$. For a judge-labeled
Evidence record, we normalize whitespace and select the last Evidence
\textsc{Add} or \textsc{Update} whose post-operation content matches its final
content. This representative-write rule may select an identical later update
that leaves the content unchanged; earlier matching writes receive no separate
credit, and differing writes are superseded. A \textsc{Discard} and its record
remain available for auditing and judging but are never selected as a positive
content write. If no matching write exists, or the selected write lacks a
mapped assistant turn or reliable browser source, the Evidence chain is dropped
rather than reassigned to an earlier write.

Candidate recovery is performed separately after semantic validation. The
operation must modify the final Candidate associated with the submission, its
normalized answer must equal the
submitted answer, and its \texttt{supporting\_clues} must contain at least one
Evidence ID whose semantic and source links both survive validation. Such an
operation advances the chain if it adds newly approved support, first creates or
revises the final answer content, or performs a genuine transition into
\texttt{metadata.status = ready}. An operation that changes none of the answer,
approved support, or readiness is a semantic no-op and is excluded.

\subsection{Source Tracing}
\label{app:source-tracing}

Each effective Evidence write is matched only to browser observations
available before the write (Appendix~\ref{app:provenance}). We first
consider the pending browser actions recorded at write time. If no
reliable match is found, we expand the candidate set to actions with
an exact recorded URL and the 80 most recent prior browser actions.

A deterministic matcher combines document IDs, URLs, search queries,
and content coverage, defined as the fraction of normalized Evidence
tokens present in an observation. Titles and recency provide auxiliary
ranking signals but cannot independently establish a reliable link.
The best-matching source is retained only if it passes fixed
source-identity or content-based reliability checks. Accepted sources
must be successful \texttt{search}, \texttt{open}, or \texttt{find}
actions, and both the source and Evidence write must map to assistant
turns. Evidence without a qualifying source is excluded from
positive attribution.

For a correct trajectory, each surviving chain connects the matched
browser action, validated state transitions, and the final submission:

\begin{equation}
\text{Source}
\longrightarrow \Delta^E
\longrightarrow [\Delta^C]_{\mathrm{optional}}
\longrightarrow \text{Final}.
\label{eq:attribution-chain}
\end{equation}
Incorrect trajectories retain only validated Source--State partial
chains within the final support set of the Candidate associated with
the submitted answer. Unfinished trajectories receive no positive
attribution. If no valid Source--State chain survives, PPO falls back
to outcome-only learning.

\section{Training and Evaluation Details}
\label{app:experimental-details}

\subsection{SFT Data and Training}
\label{app:sft-construction}
Table~\ref{tab:sft-training-config} summarizes the data and principal training
settings. Recent work explores deep-search training data construction through
grounded self-play \citep{tan2026searchmaster} and teacher-specific
entity-familiarity screening \citep{liu2026avoiding}.
ClueFlow constructs supervision from teacher
trajectories on existing QA tasks.
Both backbones receive full-parameter SFT in each environment; the
controlled SFT ablations use Qwen3-30B-A3B-Thinking-2507.

\begin{table}[H]
\centering
\footnotesize
\setlength{\tabcolsep}{4pt}
\caption{SFT data and training for the two environments.}
\label{tab:sft-training-config}
\begin{tabularx}{\textwidth}{@{}>{\raggedright\arraybackslash}p{0.33\textwidth}XX@{}}
\toprule
Setting & Fixed corpus & Live web \\
\midrule
QA source & OpenResearcher QA & REDSearcher QA \\
Training questions & 3,657 & 7,551 \\
Correct / incorrect teacher trajectories & 2,927 / 730 & 6,300 / 1,251 \\
SFT examples (one-turn / multi-target chunk) & 34,783 / 44,031 & 38,962 / 75,557 \\
Chunk size $K$ & 5 & 10 \\
Global batch size & 64 & 64 \\
Maximum sequence length & 131,072 & 163,840 \\
\midrule
\multicolumn{3}{@{}l}{\textit{Shared optimization}} \\
Epochs / learning rate & \multicolumn{2}{l@{}}{3 / $1\times10^{-5}$} \\
Schedule / warmup & \multicolumn{2}{l@{}}{Cosine decay / 0.1 warmup ratio} \\
\bottomrule
\end{tabularx}
\end{table}

\paragraph{Supervision selection.}
In correct trajectories, assistant responses are supervised unless they contain
state operations judged harmful after execution. In incorrect trajectories,
only validated Source or State turns without harmful operations are supervised;
incorrect final submissions and other turns are masked. When a turn combines
validated and harmful operations, its entire response is masked rather than
supervising individual tool calls. Masked responses may still appear in later
recorded contexts.

\paragraph{Chunk construction and masking.}
These counts are complementary parts of one hybrid export, not two encodings
of the same turns.
We group successive assistant turns into non-overlapping multi-target chunks
of at most $K$ turns using the last turn's recorded input. Only supervised,
non-attributed assistant responses receive loss within a chunk; attributed
turns use separate one-turn examples in their original contexts. Other
responses and context tokens are masked. Fixed-corpus and live-web SFT retain
10 and 20 visible turns, respectively.

\subsection{PPO Data and Training}
\label{app:ppo-training}
Table~\ref{tab:ppo-training-config} summarizes the PPO training settings for
both backbones.

\begin{table}[H]
\centering
\footnotesize
\setlength{\tabcolsep}{4pt}
\renewcommand{\arraystretch}{1.4}
\caption{PPO data and training.}
\label{tab:ppo-training-config}
\begin{tabularx}{\textwidth}{@{}>{\raggedright\arraybackslash}p{0.30\textwidth}*{2}{>{\raggedright\arraybackslash}X}@{}}
\toprule
Setting & Fixed 30B / 35B & Live 30B / 35B \\
\midrule
QA source & OpenResearcher QA & REDSearcher QA \\
Training questions sampled & 800 & 1,000 \\
Trajectories per rollout batch & 20 & 100 \\
Rollout batches & 40 & 10 \\
Optimization batch size & 20 & 100 \\
History window (turns) & 10 & 20 \\
\midrule
\multicolumn{3}{@{}l}{\textit{Shared configuration}} \\
Samples per rollout prompt & \multicolumn{2}{l@{}}{1} \\
PPO epochs per rollout batch & \multicolumn{2}{l@{}}{2} \\
Actor / critic learning rate & \multicolumn{2}{l@{}}{$10^{-6}$ / $10^{-5}$} \\
Policy / value clip & \multicolumn{2}{l@{}}{0.1 / 0.2} \\
GAE $\gamma$ / $\lambda$ & \multicolumn{2}{l@{}}{1 / 0.995} \\
Temperature / top-$p$ & \multicolumn{2}{l@{}}{1.0 / 0.95} \\
\bottomrule
\end{tabularx}
\end{table}

The outcome reward and Source/State/Final advantage bonuses are defined in
Section~\ref{sec:clueflow-ppo} (Equation~\ref{eq:chain-bonus}). Advantage
normalization is disabled, preserving the scale of these fixed bonuses. The
critic bounds its values to
$[-0.1,1]$ and applies loss only at the pre-response boundary of each valid
assistant turn, against the unmodified turn-level return.

\FloatBarrier
\subsection{Evaluation Protocols}
\label{app:evaluation-protocols}

We evaluate on all 830 BrowseComp-Plus (BCP), 1,266 BrowseComp (BC), 289
BrowseComp-ZH (BC-ZH), and 100 xBench-DeepSearch-2510 questions. Offline
training uses the OpenResearcher\footnote{\url{https://huggingface.co/collections/TIGER-Lab/openresearcher}}
corpus and released document embeddings, indexed locally with FAISS IVF. BCP
evaluation uses the BrowseComp-Plus\footnote{\url{https://huggingface.co/collections/Tevatron/browsecomp-plus}}
corpus and precomputed embeddings released by Tevatron. Both use
Qwen3-Embedding-8B for retrieval.
Live-web benchmarks use Serper search and Jina Reader for page
fetching.\footnote{Serper: \url{https://serper.dev/}; Jina Reader:
\url{https://github.com/jina-ai/reader}.}
Evaluations use a maximum of 800 browsing calls per question.
xBench-DeepSearch-2510 scores are averaged over six evaluation runs;
all other benchmarks are evaluated once.

\Needspace{7\baselineskip}
\paragraph{Inference-time context management.}
For the GLM-5.1 BrowseComp comparison in
Section~\ref{sec:inference-time}, Discard-all drops the visible trajectory
once it exceeds 64K tokens. Keep-$k$ uses $k=5$, retaining the five most
recent rounds' observations while folding older ones. HCM applies the same
keep-5 folding and discards the tool-call history if the folded context still
exceeds 64K tokens. ClueFlow retains the most recent $w=20$ turns alongside
its persistent research state. The same 64K threshold is used by HCM and
Discard-all on BCP in Table~\ref{tab:additional-context-baselines}.

The gold-conditioned GPT-5.5 attribution judge validates Evidence and
Candidate records for training and PPO credit, not benchmark accuracy. For
final-answer scoring, normalized matches are accepted first; otherwise an
answer-only GPT-5.5 judge compares the submitted answer with the reference
without viewing the search trace. Unanswered or unresolved questions count
as incorrect.

\section{Additional Results and Analyses}
\label{app:additional-results}

\subsection{Additional Context-Management Baselines}
\label{app:additional-context-baselines}

We evaluate additional context-management baselines on the full
BrowseComp-Plus evaluation set. Four inference-time strategies reuse the same
ReAct SFT model. We separately train Keep-5 SFT on teacher trajectories
collected under Keep-$k$ ($k=5$) and evaluate it under Keep-5 and HCM.
Table~\ref{tab:additional-sft-data} compares their training data with
ClueFlow SFT.

\begin{table}[htbp]
\centering
\footnotesize
\setlength{\tabcolsep}{4pt}
\caption{SFT data and trajectory construction for the context-management comparison.}
\label{tab:additional-sft-data}
\begin{tabularx}{\textwidth}{@{}>{\raggedright\arraybackslash}p{0.34\textwidth}*{3}{>{\centering\arraybackslash}X}@{}}
\toprule
Training-data property & ReAct SFT & Keep-5 SFT & ClueFlow SFT \\
\midrule
Teacher rollout configuration & ReAct & Keep-5 & ClueFlow \\
Distinct training questions & 2,927 & 2,927 & 3,657 \\
Successful teacher trajectories & 2,927 & 2,927 & 2,927 \\
Incorrect trajectories used & 0 & 0 & 730$^{\dagger}$ \\
SFT serialization & Full trajectory & One-turn & One-turn + chunks \\
Serialized training examples & 2,927 & 117,877 & 78,814 \\
Attribution-guided construction & No & No & Yes \\
\bottomrule
\end{tabularx}
\smallskip
\begin{minipage}{\textwidth}
\raggedright\footnotesize
\emph{Notes.} All three use Qwen3-30B-A3B-Thinking-2507, a GLM-5.1 teacher,
and OpenResearcher QA with offline retrieval.
$^{\dagger}$ Of ClueFlow's 730 incorrect trajectories,
only validated intermediate steps, not complete trajectories, are used.
Keep-5 one-turn examples retain the teacher-visible folded
context. Serialized-example counts do not directly measure training tokens.
\end{minipage}
\end{table}

\begin{table}[htbp]
\centering
\small
\setlength{\tabcolsep}{5pt}
\caption{Additional context-management baselines on BrowseComp-Plus. The four
ReAct rows share the same SFT checkpoint and differ only in inference-time
context management. HCM and Discard-all use a 64K-token context-management
threshold. All evaluations use a maximum budget of 800 browsing calls per
question.}
\label{tab:additional-context-baselines}
\begin{tabular*}{\textwidth}{@{\extracolsep{\fill}}llr@{}}
\toprule
SFT model & Inference-time context strategy & Acc. (\%) \\
\midrule
ReAct & Full history & 28.55 \\
ReAct & Keep-5 & 29.64 \\
ReAct & HCM & 35.06 \\
ReAct & Discard-all & 41.81 \\
Keep-5 & Keep-5 & 32.89 \\
Keep-5 & HCM & 37.11 \\
ClueFlow w/o failure recovery & Persistent state + bounded window & 46.27 \\
ClueFlow & Persistent state + bounded window & \textbf{48.31} \\
\bottomrule
\end{tabular*}
\end{table}

Context management substantially improves the ReAct SFT baseline. Discard-all
achieves the strongest result among the tested ReAct configurations, increasing
accuracy from 28.55\% to 41.81\% (+13.26~pp). Keep-5 SFT exceeds
ReAct SFT under the corresponding Keep-5 and HCM inference strategies by 3.25~pp
and 2.05~pp, respectively. Because their teacher trajectories
and training questions differ, this comparison does not isolate the effect of
bounded-context training.

Using only successful teacher trajectories, ClueFlow SFT without failure
recovery reaches 46.27\%, 4.46~pp above the strongest tested
non-ClueFlow baseline, ReAct SFT with Discard-all (41.81\%). Adding validated
intermediate steps from incorrect trajectories raises accuracy by 2.04~pp
to 48.31\% (6.50~pp above the same baseline). Thus, context management
narrows the gap between full-history ReAct SFT and ClueFlow SFT, but the tested
alternatives do not fully close it. These comparisons assess complete training
and execution configurations; the
controlled SFT ablations in Section~\ref{sec:post-training} separately examine
the contribution of attribution-guided context selection.

The relative performance of HCM and Discard-all is setting-dependent: HCM
performs better with GLM-5.1 on BrowseComp, whereas Discard-all is stronger for
the Qwen3-30B SFT policy on BrowseComp-Plus.

\subsection{PPO Rollout Behavior}
\label{app:ppo-behavior}

We report rollout statistics and post-hoc attribution diagnostics for the PPO
ablations in Table~\ref{tab:learning-ablation}(b).

\begin{table}[htbp]
\centering
\small
\setlength{\tabcolsep}{4pt}
\caption{Additional rollout, behavioral, and post-hoc attribution statistics
for the BrowseComp-Plus PPO ablations.}
\label{tab:ppo-additional-analysis}
\textbf{(a) Rollout statistics}
\vspace{2pt}

\begin{tabular*}{\textwidth}{@{\extracolsep{\fill}}lrrr@{}}
\toprule
& \multicolumn{2}{c}{Avg. browse calls} & \\
\cmidrule(lr){2-3}
Variant & Correct & Non-correct & Unfinished (\#) \\
\midrule
Outcome-only & 82.21 & 319.69 & 20 \\
Source-only attribution & 106.91 & 415.37 & 64 \\
Source+Final attribution & 52.37 & 212.04 & 19 \\
Full-chain attribution & 88.76 & 396.30 & 24 \\
+ Failure recovery & 87.26 & 428.76 & 15 \\
\bottomrule
\end{tabular*}

\vspace{7pt}
\textbf{(b) Behavioral and attribution diagnostics}
\vspace{2pt}

\begin{tabular*}{\textwidth}{@{\extracolsep{\fill}}lrrrrr@{}}
\toprule
Variant & Inspect & Raw State & Valid State & State conv. & S--S Cov. \\
& (\%) & / traj & / traj & (\%) & (\%) \\
\midrule
Outcome-only & 16.01 & 4.743 & 1.871 & 39.45 & 55.54 \\
Source-only attribution & 15.93 & 3.908 & 1.636 & 41.86 & 48.43 \\
Source+Final attribution & 13.42 & 3.600 & 1.573 & 43.71 & 50.48 \\
Full-chain attribution & 23.25 & 4.367 & 1.933 & 44.25 & 57.71 \\
+ Failure recovery & 27.09 & 4.301 & 1.904 & 44.26 & 59.04 \\
\bottomrule
\end{tabular*}
\vspace{2pt}

\begin{minipage}{\textwidth}
\scriptsize
In (a), Non-correct includes incorrect-answer and unfinished trajectories;
Unfinished counts the unfinished subset.
In (b), Inspect denotes the fraction of Browser
calls that are \texttt{open} or \texttt{find}; Raw State counts state-update
turns before validation, while Valid State counts turns retained by strict
post-execution attribution. State conv. is the fraction of raw State turns
retained as validated State progress. S--S Cov. is the percentage of evaluation
trajectories containing at least one strictly valid Source--State chain.
All variants use the same 800-call browsing budget. Post-hoc attribution
is used only for analysis and does not affect inference or benchmark scoring.
\end{minipage}
\end{table}

Full-chain attribution changes more than the frequency of browser actions.
Relative to Source-only attribution, it increases both the fraction of
inspection actions and validated State progress, with Source--State coverage
rising from 48.43\% to 57.71\%. This is not simply due to more state writing:
relative to outcome-only PPO, Full-chain produces fewer raw State-update turns
(4.367 vs.\ 4.743) but more validated State turns (1.933 vs.\ 1.871),
increasing State conversion from 39.45\% to 44.25\%. Failure recovery further
increases inspection share, while its validated State statistics remain
similar to Full-chain attribution.

Source+Final yields fewer unfinished trajectories than Source-only (19 vs.\ 64),
consistent with the intended role of Final credit in encouraging answer
submission, but does not improve accuracy. Compared with full-chain attribution,
it produces fewer validated State turns (1.573 vs.\ 1.933 per trajectory) and
lower Source--State coverage (50.48\% vs.\ 57.71\%), despite similar State
conversion (43.71\% vs.\ 44.25\%). Full-chain achieves higher accuracy with a
similarly low unfinished count (24), indicating that its advantage extends
beyond answer completion.

\FloatBarrier

\definecolor{casepanelbg}{RGB}{249,250,251}
\definecolor{casepanelborder}{RGB}{211,216,222}
\definecolor{casehint}{RGB}{74,99,116}
\definecolor{casesource}{RGB}{250,239,203}
\newmdenv[backgroundcolor=casepanelbg,linecolor=casepanelborder,
  linewidth=0.4pt,skipabove=4pt,skipbelow=4pt,innerleftmargin=8pt,
  innerrightmargin=8pt,innertopmargin=6pt,innerbottommargin=6pt,
  splittopskip=6pt,splitbottomskip=6pt]{casebody}
\newenvironment{caseblock}{%
  \par\noindent\begin{minipage}{\linewidth}%
}{%
  \end{minipage}\par
}
\lstdefinestyle{casemessage}{
  basicstyle=\ttfamily\footnotesize,breaklines=true,breakatwhitespace=true,
  columns=fullflexible,keepspaces=true,showstringspaces=false,
  aboveskip=1pt,belowskip=1pt,escapeinside={(*@}{@*)}
}
\newcommand{\caseturnheading}[3]{%
  \Needspace{9\baselineskip}%
  \par\smallskip\noindent\textcolor{casepanelborder}{\rule{\linewidth}{0.4pt}}\par\nobreak
  \noindent\textbf{Turn #1: #2}\hfill
  {\footnotesize\textsf{\textcolor{casehint}{#3}}}\par\nobreak
}

\Needspace{18\baselineskip}
\section{Qualitative Case Studies}
\label{app:qualitative}

Turn numbers, tool calls, observations, and quoted Think excerpts come from
saved trajectories; ellipses mark omissions and line wraps are editorial.
The turn-8 input snapshot abbreviates a logged request while preserving role
order. Task summaries, state annotations, highlights, and Source/State/Final
labels are post-hoc, not agent-visible. Bonuses are PPO actor advantage bonuses
(Equation~\ref{eq:chain-bonus}), not environment rewards.

\Needspace{12\baselineskip}
\noindent\textbf{Correct submission.}\par\nobreak
\begin{casebody}\small
\noindent\fcolorbox{casepanelborder}{white}{\parbox{\dimexpr\linewidth-2\fboxsep-2\fboxrule\relax}{%
\textbf{Task summary.} Name the annual two-day music and
hot-air-balloon festival in Clark Global City, a planned district in a Central
Luzon economic zone; its name evokes a natural phenomenon.\par
\textbf{Reference answer.} Aurora Music Festival.\quad
\textbf{Submitted answer.} Aurora Music Festival.}}\par

\begin{caseblock}
\caseturnheading{4}{Competing event}{Unattributed}
\noindent\textbf{Think.} \emph{``Let me check if there's an event
specifically at Clark Global City.''}\par
\noindent\textsf{Assistant tool calls}\par\nobreak
\begin{lstlisting}[style=casemessage]
browser.open({"id":12})
browser.search({"query":"Lubao International Balloon Music
  Festival art installations","topn":10})
\end{lstlisting}
\end{caseblock}
\begin{caseblock}
\noindent\textsf{Tool observations}\par\nobreak
\begin{lstlisting}[style=casemessage]
browser.open: [2] Philippine International Hot Air Balloon Fiesta
  L22: ... an annual four-day air-sporting event ...
browser.search: [33] Lubao International Balloon and Music Festival 2026 ...
\end{lstlisting}
\end{caseblock}

\begin{caseblock}
\caseturnheading{6}{Source page}{Source}
\noindent\textbf{Think.} \emph{``Let me check the Wikipedia page for
the Aurora Music Festival.''}\par
\noindent\textsf{Assistant tool call:} \lstinline!browser.open({"id":43})!\par
\noindent\textsf{Tool observation}\par\nobreak
\begin{lstlisting}[style=casemessage]
[3] ## First edition and COVID-19 pandemic (73 lines total)
L1: Title: Aurora Music Festival
L3: URL Source: https://en.wikipedia.org/wiki/Aurora_Music_Festival
L24: ... art installations, and (*@\colorbox{casesource}{\texttt{hot air balloons}}@*) held in
     (*@\colorbox{casesource}{\texttt{Clark Global City}}@*) ... over the course of (*@\colorbox{casesource}{\texttt{two days.}}@*)
\end{lstlisting}
\end{caseblock}

\begin{caseblock}
\caseturnheading{7}{Record three Evidence clues}{State}
\noindent\textbf{Think.} \emph{``The Wikipedia page for the Aurora
Music Festival explicitly says it's held in Clark Global City. \ldots\ Let
me add clues and create a candidate.''}\par
\noindent\textsf{Assistant tool calls}\par\nobreak
\begin{lstlisting}[style=casemessage]
clue.add({"kind":"evidence","content":"The Aurora Music Festival
  ... held in Clark Global City ... two days. [Wikipedia]",
  "reason":"Matches all key constraints ..."})
clue.add({"kind":"evidence","content":"Clark Global City is a planned
  central business district ... [Search results 18, 23]", ...})
clue.add({"kind":"evidence","content":"Pampanga province was created
  in 1571 ... [Wikipedia/Doc 49-52]", ...})
\end{lstlisting}
\end{caseblock}
\begin{caseblock}
\noindent\textsf{Tool observations}\par\nobreak
\begin{lstlisting}[style=casemessage]
ok: added c1 (1 active); ok: added c2 (2 active);
ok: added c3 (3 active)
\end{lstlisting}
\end{caseblock}
\begin{caseblock}
\noindent\textbf{State after execution.}
The complete \texttt{c1} Evidence record is:\par\nobreak
\begin{lstlisting}[style=casemessage]
{"id":"c1","kind":"evidence","metadata":{},"status":"active",
 "content":"The Aurora Music Festival (also called the
 \"Clark Hot Air Balloon Music, Arts and Food Festival\") is
 an annual music and arts festival featuring live music, food,
 art installations, and hot air balloons held in Clark Global
 City, Pampanga, Philippines over the course of two days.
 [Wikipedia]"}
\end{lstlisting}
\noindent\texttt{c2} records Clark Global City, the SEZ, and the Bank of China
loan; \texttt{c3} records Pampanga's 1571 establishment. Both remain active.
\end{caseblock}

\begin{caseblock}
\caseturnheading{8}{Create the Candidate}{State}
\noindent\textbf{Model input.}
The rendered state precedes the recent assistant/tool history:\par\nobreak
\begin{lstlisting}[style=casemessage]
system: You are a clue-based search agent. ...
user: In a Southeast Asian nation ... two-day event ... name?
system: [Evidence Clues] c1: Aurora Music Festival ...
        c2: Clark Global City ...; c3: Pampanga ... 1571 ...
        [Candidate State] (no candidate clues)
assistant (t6): browser.open({"id":43})
tool (browser.open): [3] Aurora Music Festival ... L24: ... two days.
assistant (t7): clue.add(...) x 3
tool (clue.add): ok: added c1; ok: added c2; ok: added c3
\end{lstlisting}
\end{caseblock}
\begin{caseblock}
\noindent\textbf{Think.} \emph{``Now let me add the candidate clue
with the answer.''}\par
\noindent\textsf{Assistant tool call}\par\nobreak
\begin{lstlisting}[style=casemessage]
clue.add({"content":"Aurora Music Festival","kind":"candidate",
  "metadata":{"answer":"Aurora Music Festival","status":"ready",
  "supporting_clues":["c1","c2","c3"], ...}})
\end{lstlisting}
\end{caseblock}
\begin{caseblock}
\noindent\textsf{Tool observation}\par\nobreak
\begin{lstlisting}[style=casemessage]
ok: added c4 (4 active)
\end{lstlisting}
\end{caseblock}
\begin{caseblock}
\noindent\textbf{State after execution.}
\texttt{c4}:\par\nobreak
\begin{lstlisting}[style=casemessage]
{"id":"c4","content":"Aurora Music Festival","kind":"candidate",
 "status":"active",
 "metadata":{"answer":"Aurora Music Festival","status":"active",
  "supporting_clues":["c1","c2","c3"],
  "satisfied_constraints":["Southeast Asian nation (Philippines)",
    ...],
  "missing_constraints":[],"contradictions":[],"next_check":""}}
\end{lstlisting}
\noindent The top-level \texttt{status} tracks record lifecycle;
\texttt{metadata.status} tracks Candidate readiness. The call requested
\texttt{ready}, but creation initialized \texttt{metadata.status=active}.
\end{caseblock}

\begin{caseblock}
\caseturnheading{10}{Update readiness}{State}
\noindent\textbf{Think.} \emph{``The candidate has all constraints
satisfied, no missing constraints, no contradictions, and no next check needed.
\ldots\ Let me try updating the candidate with status ready.''}\par
\noindent\textsf{Assistant tool call}\par\nobreak
\begin{lstlisting}[style=casemessage]
clue.update({"id":"c4","metadata":{"answer":"Aurora Music Festival",
  "status":"ready","supporting_clues":["c1","c2","c3"], ...}})
\end{lstlisting}
\end{caseblock}
\begin{caseblock}
\noindent\textsf{Tool observation}\par\nobreak
\begin{lstlisting}[style=casemessage]
ok: updated c4 (4 active)
\end{lstlisting}
\noindent\textbf{State change.}
\texttt{c4.metadata.status}: \texttt{active} $\to$ \texttt{ready};
\texttt{c4.status} stays \texttt{active}. The answer and
\texttt{supporting\_clues=[c1,c2,c3]} do not change.
\end{caseblock}

\begin{caseblock}
\caseturnheading{11}{Submit the answer}{Final}
\noindent\textbf{Think.} \emph{``I should answer with the plain text
answer.''}\par
\noindent\textsf{Assistant final answer:}
\texttt{Aurora Music Festival}
\end{caseblock}
\end{casebody}

\Needspace{11\baselineskip}
\noindent{\raggedright\textit{Post-hoc attribution.} Validated \texttt{c1}
(\texttt{SUPPORT\_GOLD}) links turn 6 to turn 7 and supports \texttt{c4};
\texttt{c2}/\texttt{c3} supply context without independent credited chains.\par}
\noindent{\footnotesize\begin{tabularx}{\linewidth}{@{}l l X r@{}}
\toprule
Turn & Role & Reason & Actor advantage bonus \\
\midrule
4 & Unattributed & Competing festival and exploratory search & $0$ \\
6 & Source & Matched page for \texttt{c1} & $+0.03$ \\
7 & State & Validated \texttt{c1} write (one turn, three \texttt{clue.add} calls) & $+0.01$ \\
8 & State & Validated Candidate creation with \texttt{c1} support & $+0.01$ \\
10 & State & \texttt{c4.metadata.status}: \texttt{active}$\to$\texttt{ready} & $+0.01$ \\
11 & Final & Correct answer submission & $+0.04$ \\
\bottomrule
\end{tabularx}}

\Needspace{12\baselineskip}
\noindent\textbf{Incorrect submission.}\par\nobreak
\begin{casebody}\small
\noindent\fcolorbox{casepanelborder}{white}{\parbox{\dimexpr\linewidth-2\fboxsep-2\fboxrule\relax}{%
\textbf{Task summary.} Which specific Pakistani TV, fashion, and
music awards ceremony was held at a major Dubai convention venue? The full
question adds historical-government and public-figure clues.\par
\textbf{Reference answer.} 3rd Hum Awards.\quad
\textbf{Submitted answer.} Hum Awards.}}\par

\begin{caseblock}
\caseturnheading{13}{Unrelated find}{Unattributed}
\noindent\textbf{Think.} \emph{``Let me look at the notable residents
section.''}\par
\noindent\textsf{Assistant tool call}\par\nobreak
\begin{lstlisting}[style=casemessage]
browser.find({"pattern":"Notable past residents",
  "cursor":1,"context_lines":5})
\end{lstlisting}
\end{caseblock}
\begin{caseblock}
\noindent\textsf{Tool observation}\par\nobreak
\begin{lstlisting}[style=casemessage]
Found 2 match(es) for 'Notable past residents' in [1]:
L230: ... Aqeela Asifi ... Nansen Refugee Award ...
\end{lstlisting}
\end{caseblock}

\begin{caseblock}
\caseturnheading{18}{Awards page}{Source}
\noindent\textbf{Think.} \emph{``Actually, let me do one more
verification. Let me check the specific name of the ceremony.''}\par
\noindent\textsf{Assistant tool call}\par\nobreak
\begin{lstlisting}[style=casemessage]
browser.open({"id":6,"loc":0,"num_lines":50})
\end{lstlisting}
\end{caseblock}
\begin{caseblock}
\noindent\textsf{Tool observation}\par\nobreak
\begin{lstlisting}[style=casemessage]
[2] Hum Awards (81 lines total) [showing L1-L50]
L1: Title: Hum Awards
L15: Country: Pakistan
L22: Hum Awards are annual Pakistani entertainment awards
     ... Pakistani television, fashion and music ...
L28: ... beginning with the (*@\colorbox{casesource}{\texttt{3rd Hum Awards}}@*) in Dubai in 2015.
L40: | (*@\colorbox{casesource}{\texttt{3rd Hum Awards}}@*) | 9 April 2015 | ... |
     Dubai World Trade Centre
\end{lstlisting}
\end{caseblock}

\begin{caseblock}
\caseturnheading{19}{Write valid Evidence}{State}
\noindent\textbf{Think.} \emph{``The remaining parts of the chain
(three historical polities, Afghanistan public figure) \ldots\ but they're
not strictly necessary to answer the question \ldots''}\par
\noindent\textsf{Assistant tool call}\par\nobreak
\begin{lstlisting}[style=casemessage]
clue.add({"kind":"evidence","content":"Hum Awards match all
  core criteria ... The 3rd Hum Awards was held at Dubai
  World Trade Centre in 2015.","reason":"Confirms all three
  core identifying features of the ceremony ..."})
\end{lstlisting}
\end{caseblock}
\begin{caseblock}
\noindent\textsf{Tool observation}\par\nobreak
\begin{lstlisting}[style=casemessage]
ok: added c3 (3 active)
\end{lstlisting}
\end{caseblock}
\begin{caseblock}
\noindent\textbf{State after execution.}
The \texttt{c3} Evidence record has these fields:\par\nobreak
\begin{lstlisting}[style=casemessage]
{"id":"c3","kind":"evidence","metadata":{},"status":"active",
 "content":"Hum Awards match all core criteria ... The
 (*@\colorbox{casesource}{\texttt{3rd Hum Awards}}@*) was held at Dubai World Trade Centre in 2015."}
\end{lstlisting}
\end{caseblock}

\begin{caseblock}
\caseturnheading{20}{Create an incorrect Candidate}{Unattributed}
\noindent\textbf{Think.} \emph{``Now let me create a candidate clue
with the answer.''}\par
\noindent\textsf{Assistant tool call}\par\nobreak
\begin{lstlisting}[style=casemessage]
clue.add({"content":"Hum Awards","kind":"candidate",
  "metadata":{"answer":"Hum Awards","status":"ready",
  "supporting_clues":["c1","c2","c3"],
  "missing_constraints":["Verifying Afghanistan public figure
  connection","Verifying three historical polities ..."], ...}})
\end{lstlisting}
\end{caseblock}
\begin{caseblock}
\noindent\textsf{Tool observation}\par\nobreak
\begin{lstlisting}[style=casemessage]
ok: added c4 (4 active)
\end{lstlisting}
\end{caseblock}
\begin{caseblock}
\noindent\textbf{State after execution.}
{\raggedright\texttt{c4} has \texttt{kind=candidate},
\texttt{content="Hum Awards"}, top-level \texttt{status=active},
\texttt{metadata.answer="Hum Awards"}, \texttt{metadata.status=active}
(despite requested \texttt{ready}), and \texttt{supporting\_clues=[c1,c2,c3]}.
Two constraints remain missing. Evidence \texttt{c3} already names the
\emph{3rd} edition, but \texttt{c4} does not.\par}
\end{caseblock}

\begin{caseblock}
\caseturnheading{36}{Update readiness}{Unattributed}
\noindent\textbf{Think.} \emph{``So I'm confident the answer is the
Hum Awards. Let me update the candidate and mark it as ready.''}\par
\noindent\textsf{Assistant tool call}\par\nobreak
\begin{lstlisting}[style=casemessage]
clue.update({"id":"c4","metadata":{"answer":"Hum Awards",
  "status":"ready","missing_constraints":[], ...}})
\end{lstlisting}
\noindent\textsf{Tool observation}\par\nobreak
\begin{lstlisting}[style=casemessage]
ok: updated c4 (4 active)
\end{lstlisting}
\noindent\textbf{State change.}
\texttt{c4.metadata.status}: \texttt{active} $\to$ \texttt{ready};
\texttt{missing\_constraints}: two checks $\to$ \texttt{[]}.
\texttt{metadata.answer="Hum Awards"} and
\texttt{supporting\_clues=[c1,c2,c3]} remain unchanged.
\end{caseblock}

\begin{caseblock}
\caseturnheading{37}{Submit an incorrect answer}{Unattributed}
\noindent\textbf{Think.} \emph{``Let me provide the final answer.''}\par
\noindent\textsf{Assistant final answer:} \texttt{Hum Awards}
\end{caseblock}
\end{casebody}

\Needspace{11\baselineskip}
\noindent{\raggedright\textit{Post-hoc attribution.} Evidence \texttt{c3} is
\texttt{SUPPORT\_GOLD}, links to turn 18, and is cited in the submitted
Candidate's \texttt{supporting\_clues}. It retains a Source--State partial
chain even though \texttt{c4} and the answer omit the edition, complementing
prior work on failure localization and repair in
long-horizon search \citep{liang2026searchauditor}. Incorrect
Candidate operations and the final submission receive no positive credit.\par}
\noindent{\footnotesize\begin{tabularx}{\linewidth}{@{}l l X r@{}}
\toprule
Turn & Role & Reason & Actor advantage bonus \\
\midrule
13 & Unattributed & Unrelated \texttt{find} & $0$ \\
18 & Source & Matched awards page for \texttt{c3} & $+0.03$ \\
19 & State & Validated \texttt{c3} write naming the 3rd edition & $+0.03$ \\
20 & Unattributed & Candidate answer remains \emph{Hum Awards} & $0$ \\
36 & Unattributed & Marks that answer \texttt{ready}, still no edition & $0$ \\
37 & No Final credit & Incorrect answer submission & $0$ \\
\bottomrule
\end{tabularx}}

\Needspace{9\baselineskip}
\section{Limitations and Future Work}
\label{app:limitations}

The present study instantiates state-transition attribution for long-horizon
search using Evidence and Candidate records. Generalizing this learning
interface to broader agent settings calls for state abstractions and
progress-validation criteria tailored to different tasks. Future work will
explore extensions to open-ended research and multimodal interaction,
examining how attributed state transitions can support learning beyond
answer-oriented search.

\end{document}